\documentclass[11pt,a4paper]{article}
\usepackage[hyperref]{emnlp-ijcnlp-2019}
\aclfinalcopy
\usepackage{times}
\usepackage{latexsym}
\usepackage{graphicx}
\usepackage{amsmath}
\usepackage{hyperref}
\usepackage{url}
\allowdisplaybreaks
\usepackage{tikz}
\usetikzlibrary{arrows}
\usepackage{xspace}
\usepackage{latexsym}
\usepackage{amsfonts}
\usepackage{amsmath}
\usepackage{amssymb}
\usepackage{color}
\usepackage{colortbl}
\usepackage{epsfig}
\usepackage{graphicx}
\usepackage{pbox}
\usepackage{paralist}
\usepackage{enumerate}
\usepackage[color,matrix,arrow,all]{xy}
\usepackage{comment}
\usepackage{booktabs}
\usepackage{balance}
\usepackage{stmaryrd}
\usepackage{pifont}
\usepackage{hhline}
\usepackage{listings}
\usepackage{array}
\usepackage{float}
\usepackage[flushleft]{threeparttable}
\usepackage{graphicx,wrapfig,lipsum}
\usepackage{subcaption}
\usetikzlibrary{arrows}
\usepackage{graphicx,dblfloatfix}

\DeclareMathAlphabet{\pazocal}{OMS}{zplm}{m}{n}

\usepackage{array}
\newcolumntype{M}{>{\begin{varwidth}{2cm}}l<{\end{varwidth}}}
\newcolumntype{L}[1]{>{\raggedright\let\newline\\\arraybackslash\hspace{0pt}}m{#1}}
\newcolumntype{C}[1]{>{\centering\let\newline\\\arraybackslash\hspace{0pt}}m{#1}}
\newcolumntype{R}[1]{>{\raggedleft\let\newline\\\arraybackslash\hspace{0pt}}m{#1}}

\newcommand{\ie}{{\em i.e.,}\xspace}

 \usepackage[framemethod=TikZ]{mdframed}
\usetikzlibrary{shadows}
\mdfdefinestyle{myframe}{%
    linecolor=black,
    outerlinewidth=0.6pt,
    roundcorner=10pt,
    innertopmargin=10pt,
    innerbottommargin=10pt,
    innerrightmargin=10pt,
    innerleftmargin=10pt,
    skipabove=0.6\baselineskip,
    }

\newcommand{\Ls}{\mathcal{L}}

\title{Using Clinical Notes with Time Series Data for ICU Management}

\author{Swaraj Khadanga \\
  University of Minnesota \\
  Minneapolis, MN  \\
  {\tt khada004@umn.edu} \\\And
  Karan Aggarwal \\
  University of Minnesota \\
  Minneapolis, MN  \\
  {\tt aggar081@umn.edu} \\\And
  Shafiq Joty \\
  Nanyang Technological University \\
  Singapore  \\
  {\tt srjoty@ntu.edu.sg} \\\AND
  Jaideep Srivastava \\
  University of Minnesota \\
  Minneapolis, MN  \\
  {\tt srivasta@umn.edu}}

\date{}

\begin{document}
\maketitle
\begin{abstract}
Monitoring patients in ICU is a challenging and high-cost task. Hence, predicting the condition of patients during their ICU stay can help provide better acute care and plan the hospital's resources. There has been continuous progress in machine learning research for ICU management, and most of this work has focused on using time series signals recorded by ICU instruments. In our work, we show that adding clinical notes as another modality improves the performance of the model for three benchmark tasks: in-hospital mortality prediction, modeling decompensation, and length of stay forecasting that play an important role in ICU management. While the time-series data is measured at regular intervals, doctor notes are charted at irregular times, making it challenging to model them together. We propose a method to model them jointly, achieving considerable improvement across benchmark tasks over baseline time-series model. Our implementation can be found at \url{https://github.com/kaggarwal/ClinicalNotesICU}.
\end{abstract}

\section{Introduction}

With the advancement of medical technology, patients admitted into the intensive care unit (ICU) are monitored by different instruments on their bedside, which measure different vital signals about patient's health. 
During their stay, doctors visit the patient intermittently for check-ups and make \emph{clinical notes} about the patient's health and physiological progress. These notes can be perceived as \emph{summarized expert knowledge} about the patient's state. All these data about instrument readings, procedures, lab events, and clinical notes are recorded for reference. 
Availability of ICU data and enormous progress in machine learning have opened up new possibilities for health care research. 
Monitoring patients in ICU is a challenging and high-cost task. Hence, predicting the condition of patients during their ICU stay can help plan better resource usage for patients that need it most in a cost-effective way. Prior works~\cite{harutyunyan2017multitask, ghassemi2015multivariate, suresh2018learning, song2018attend,caballero2015dynamically} have focused exclusively on modeling the problem using the time series signals from medical instruments. Expert knowledge from doctor's notes has been ignored in the literature.

In this work, we use clinical notes in addition to the time-series data for improved prediction on benchmark ICU management tasks \cite{harutyunyan2017multitask}. 
While the time-series data is measured continuously, the doctor notes are charted at intermittent times. This creates a new challenge to model continuous time series and discrete-time note events jointly. We propose such a multi-modal deep neural network that comprises of recurrent units for the time-series and convolution network for the clinical notes. We demonstrate that adding clinical notes improves the performance on in-hospital mortality prediction, modeling decompensation, and length of stay forecasting tasks.

\begin{figure*}
	\centering
	\includegraphics[width=0.9\textwidth]{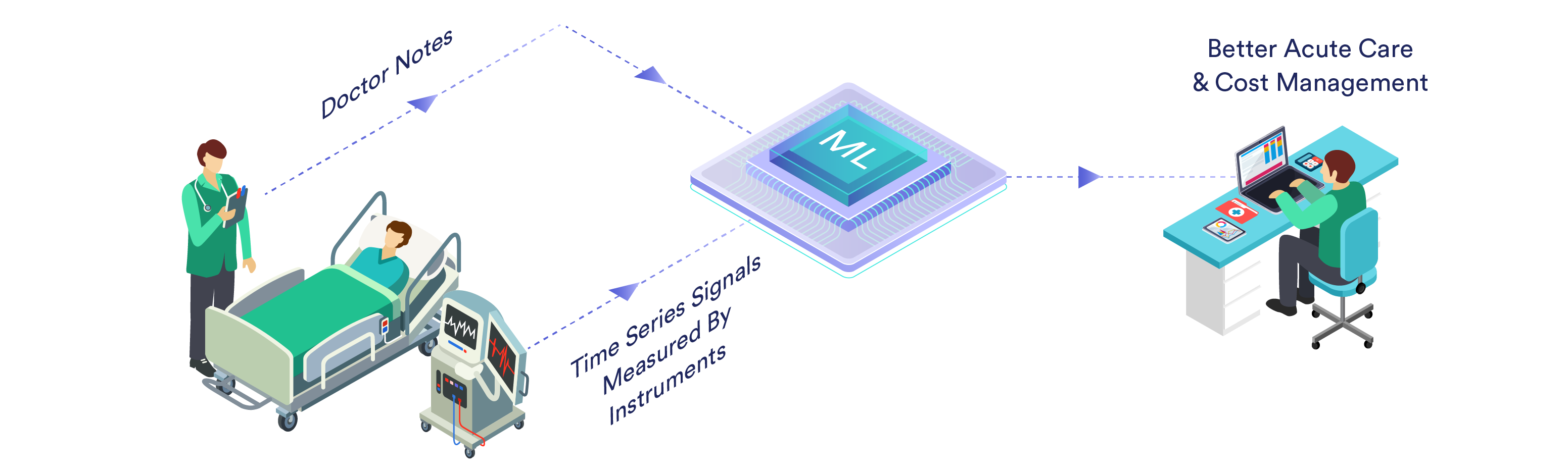}
	\caption{Doctor notes compliments measured physiological signals for better ICU management.}
	\label{fig:the-idea}
\end{figure*}

\section{Related Work}
We provide a review of machine learning approaches for clinical prediction tasks. 

\paragraph{Biomedical natural language processing}
The recent developments in deep learning-based techniques for NLP have been adapted for clinical notes. Convolutional neural networks have been used to predict ICD codes from clinical texts \cite{mullenbach-etal-2018-explainable,8320340,gangavarapu2019tags}. \newcite{rios2015convolutional} and \newcite{baker2016cancer} used convolutional neural networks to classify various biomedical articles. Pre-trained word and sentence embeddings have also shown good results for sentence similarity tasks~\cite{chen2018biosentvec}. Recently, there is an interest in the community to use clinical notes for ICU related tasks~\cite{jin2018improving,boag2018s,liu2019knowledge,huang2019clinicalbert}. Given the long structured nature of the clinical text, we prefer convolutional neural networks over recurrent networks, as demonstrated in previous studies~\cite{zhang2016rationale,boag2018s}. The work closest to ours is~\newcite{jin2018improving}, who use an aggregated word embeddings of clinical notes for in-hospital mortality prediction. 

\paragraph{ICU management related literature} ICU management literature has focused exclusively on using time-series measurements for the prediction tasks~\newcite{harutyunyan2017multitask, ghassemi2015multivariate, suresh2018learning, song2018attend,caballero2015dynamically}. Recurrent neural networks have been models of choice for these recent works, with additional gains from using attention or multi-task learning (\newcite{song2018attend}). \newcite{xu2018raim} accommodated supplemental information like diagnosis, medications, and lab events to improve model performance. We use RNNs for modeling time-series in this work, utilizing the setup identical to ~\newcite{harutyunyan2017multitask}. 

Multi-modal learning has shown success in speech, natural language, and computer vision (\newcite{ngiam2011multimodal}, \newcite{mao2014explain}). Recently, a lot of work has been done using images/videos with natural language text (\newcite{elliott2016multimodal}). We use a similar intuition for utilizing clinical notes with time-series data for ICU management tasks. In the next section, we define the three benchmark tasks we evaluate in this work.

\section{Prediction Tasks}
We use the definitions of the benchmark tasks defined by \newcite{harutyunyan2017multitask} as the following three problems:
\begin{enumerate}
\setlength\itemsep{0.2em}
    \item \textbf{In-hospital Mortality}: This is a binary classification problem to predict whether a patient dies before being discharged from the first two days of ICU data.
    \item \textbf{Decompensation}: Focus is to detect patients who are physiologically declining. 
    Decompensation is defined as a sequential prediction task where the model has to predict at each hour after ICU admission. Target at each hour is to predict the mortality of the patient within a 24 hour time window.
    \item \textbf{Length of Stay Forecasting (LOS)}: The benchmark defines LOS as a prediction of bucketed remaining ICU stay with a multi-class classification problem. Remaining ICU stay time is discretized into 10 buckets: $\{0-1,1-2,2-3,3-4,4-5,5-6,6-7,7-8,8-14,14+\}$ days where first bucket, covers the patients staying for less than a day (24 hours) in ICU and so on. This is only done for the patients that did not die in ICU. 
\end{enumerate}
These tasks have been identified as key performance indicators of models that can be beneficial in ICU management in the literature. 
Most of the recent work has focused on using RNN to model the temporal dependency of the instrument time series signals for these tasks (\newcite{harutyunyan2017multitask}, \newcite{song2018attend}).

\section{Methods}
In this section, we describe the models used in this study. We start by introducing the notations used, then describe the baseline architecture, and finally present our proposed multimodal network.

For a patient's length of ICU stay of  $T$ hours, we have time series observations, $x_t$ at each time step $t$ (1 hour interval) measured by instruments along with doctor's note $n_i$ recorded at \emph{irregular} time stamps. Formally, for each patient's ICU stay, we have time series data $[x_{t}]_{t=1}^T$ of length $T$, and $K$ doctor notes $[N_i]_{i=1}^K$ charted at time $[TC(i)]_{i=1}^K$, where $K$ is generally much smaller than $T$. For \textbf{in-hospital mortality} prediction, $m$ is a binary label at $t=48$ hours, which indicates whether the person dies in ICU before being discharged. For \textbf{decompensation} prediction performed hourly, $[d_{t}]_{t=5}^{T}$ are the binary labels at each time step $t$, which indicates whether the person dies in ICU within the next 24 hours. For \textbf{LOS} forecasting also performed hourly, $[l_t]_{t=5}^T$ are multi-class labels defined by buckets of the remaining length of stay of the patient in ICU.  Finally, we denote $N_{T}$ as the concatenated doctor's note during the ICU stay of the patient (\ie, from $t=1$ to $t=T$). 


\subsection{Baseline: Time-Series LSTM Model} \label{sec:base}
Our baseline model is similar to the models defined by \newcite{harutyunyan2017multitask}. For all the three tasks, we used a Long Short Term Memory or LSTM~\cite{hochreiter1997long} network to model the temporal dependencies between the time series observations, $[x_{t}]_{t=1}^T$. At each step, the LSTM composes the current input $x_t$ with its previous hidden state $h_{t-1}$ to generate its current hidden state $h_t$; that is, $h_t = \text{LSTM}(x_t, h_{t-1})$ for $t=1$ to $t=T$. The predictions for the three tasks are then performed with the corresponding hidden states as follows:
\begin{equation}
\begin{split}
    \hat{m} &= \text{sigmoid}(W_m h_{48}+b_m) \\
    \hat{d_{t}} &= \text{sigmoid}(W_d h_{t}+b_d) \hspace{0.5em} \text{for}~ t=5 \ldots T\\
    \hat{l_{t}} &= \text{softmax}(W_l h_{t}+b_l) \hspace{0.5em} \text{for}~ t=5 \ldots T \\
\end{split}
\end{equation}
where $\hat{m}$, $\hat{d}_t$, and $\hat{l}_t$ are the probabilities for in-hospital mortality, decompensation, and LOS, respectively, and $W_m$, $W_d$, and $W_l$ are the respective weights of the fully-connected (FC) layer. Notice that the in-hospital mortality is predicted at end of 48 hours, while the predictions for decompensation and LOS tasks are done at each time step after first four hours of ICU stay. We trained the models using cross entropy (CE) loss defined as below.
\begin{equation}
\begin{split}
    \Ls_{\mathrm{ihm}}&= \text{CE}(m,\hat{m})\\
    \Ls_{\mathrm{decom}}&= \frac{1}{T}\sum_t \text{CE}(d_{t},\hat{d_{t}})\\
    \Ls_{\mathrm{los}}&= \frac{1}{T}\sum_t \text{CE}(l_{t},\hat{l_{t}})\\
\end{split}
\end{equation}
\subsection{Multi-Modal Neural Network}

In our multimodal model, our goal is to improve the predictions by taking both the time series data $x_{t}$ and the doctor notes $n_{i}$ as input to the network. 



\paragraph{Convolutional Feature Extractor for Doctor Notes.}
As shown in Fig. \ref{fig:the-idea}, we adopt a convolutional approach similar to~\newcite{kim-2014-convolutional} to  extract the textual features from the doctor's notes. For a piece of clinical note $N$, our CNN takes the word embeddings $\mathbf{e}=(e_1,e_2,\dots,e_n)$ as input and applies 1D convolution operations, followed by max-pooling over time to generate a $p$ dimensional feature vector $\hat{z}$, which is fed to the fully connected layer along side the LSTM output from time series signal (described in the next paragraph) for further processing. From now onwards, we denote the 1D convolution over note $N$ as $\hat{z} = \text{Conv1D}(N)$. 
\begin{figure}
\centering
    \includegraphics[scale=1]{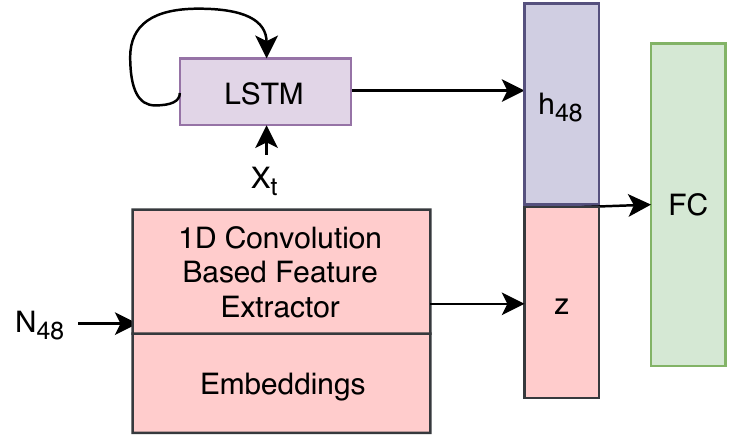}
    \caption{Block diagram from the in-hospital mortality multi-modal network.}
    \label{fig:the-idea}
\end{figure}
\paragraph{Model for In-Hospital Mortality.} This model takes the time series signals $[x_t]_{t=1}^{T}$ and all notes $[N_i]_{i=1}^{K}$ to predict the mortality label $m$ at $t=T$ ($T = 48$). For this, $[x_t]_{t=1}^{T}$ is processed through an LSTM layer just like the baseline model in Sec. \ref{sec:base}, and for the notes, we concatenate ($\otimes$) all the notes $N_{1}$ to $N_{K}$ charted between $t=1$ to $t=T$ to generate a single document $N_{T}$. More formally, 
\begin{equation}
\begin{split}
N_{T} &= N_{1} \otimes N_{2} \otimes \dots \otimes N_{K}\\
h_{t} &= \text{LSTM}(x_{t},h_{t-1}) \hspace{0.5em} \text{for}~ t=1 \ldots T \\
\hat{z} &= \text{Conv1D}(N_{T})\\
\hat{m}  &=  \text{sigmoid}(W_1h_{48} + W_2 \hat{z} + b)  \\
\end{split}
\end{equation}
We use pre-trained word2vec embeddings \cite{mikolov2013distributed} trained on both MIMIC-III clinical notes and PubMed articles to initialize our methods as it outperforms other embeddings as shown in \cite{chen2018biosentvec}. We also freeze the embedding layer parameters, as we did not observe any  improvement by fine-tuning them. 
\paragraph{Model for Decompensation and Length of Stay.} Being sequential prediction problems, modeling decompensation and length-of-stay requires special technique to align the discrete text events to continuous time series signals, measured at 1 event per hour. Unlike in-hospital mortality, here we extract feature maps $z_{i}$ by processing each note $N_{i}$ independently using 1D convolution operations. For each time step $t=1,2\dots T$, let $z_{t}$ denote the extracted text feature map to be used for prediction at time step $t$. We compute $z_{t}$ as follows.
\begin{equation}
    \begin{split}
        z_{i}  &= \text{Conv1D}(N_{i}) \hspace{0.5em} \text{for}~i=1\dots K\\
        w(t,i) &= \exp[{-\lambda*(t-CT(i))}]\\
        {z}_{t}  &= \frac{1}{M} \sum_{i=1}^M z_{i} w(t,i) \\
    \end{split}
\end{equation}

\noindent where $M$ is the number of doctor notes seen before time-step $t$, and $\lambda$ is a decay hyperparameter tuned on a validation data. Notice that ${z}_{t}$ is computed as a weighted sum of the feature vectors, where the weights are computed with an exponential decay function. The intuition behind using a decay is to give preference to recent notes as they better describe the current state of the patient. 

The time series data $x_{t}$ is modeled using an LSTM as before. We concatenate the attenuated output from the CNN with the LSTM output for the prediction tasks as follows:
\begin{equation}
    \begin{split}
        h_{t}  &= \text{LSTM}(x_{t},h_{t-1}) \\
        \hat{d_{t}}  &=  \text{sigmoid}(W_d^1 h_{t} + W_d^2{z}_{t} + b)  \\
        \hat{l}_{t}  &=  \text{softmax}(W_l^1h_{t} + W_l^2{z}_{t} + b)  \\
    \end{split}
\end{equation}

\begin{figure}
    \includegraphics[width=\columnwidth]{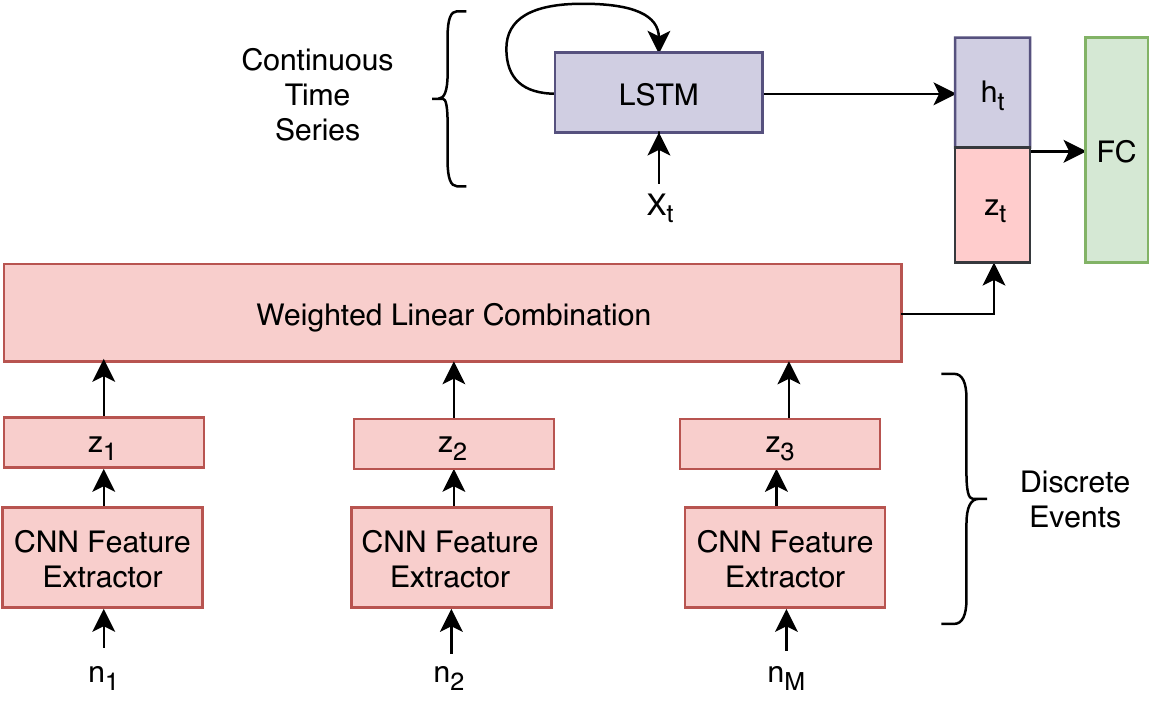}
    \caption{Block diagram from decompensation and length of stay prediction multi-modal network.}
    \label{fig:the-idea2}
\end{figure}

\noindent Both our baselines and multimodal networks are regularized using dropout and weight decay. We used Adam Optimizer to train all our models.

\section{Experiments}
We used MIMIC-III \cite{johnson2016mimic} dataset for all our experiments 
following \newcite{harutyunyan2017multitask}'s benchmark setup for processing the time series signals from ICU instruments. We use the same test-set defined in the benchmark and 15\% of remaining data as validation set. For in-hospital mortality task, only those patients are considered who were admitted in the ICU for at least 48 hours. 
However, we dropped all clinical notes which doesn't have any chart time associated and also dropped all the patients without any notes. Owing to this step, \textit{our results are not directly comparable to the numbers reported by \newcite{harutyunyan2017multitask}}.  
Notes which have been charted before ICU admission are concatenated and treated as one note at $t=1$. After pre-processing, the number of patients for in-hospital mortality is 11,579 and 22,353 for the other two tasks. 

For in-hospital mortality task, best performing baseline and multimodal network have 256 hidden units LSTM cell. For convolution operation, we used 256 filters for each of kernel size 2, 3 and 4. For decompensation and LOS prediction, we used 64 hidden units for LSTM and 128 filters for each 2,3 and 4 size convolution filters. The best decay factor $\lambda$ for text features was 0.01. We implement our methods with TensorFlow~\newcite{tensorflow2015-whitepaper}\footnote{The code can be found at \url{https://github.com/kaggarwal/ClinicalNotesICU}}. All our models were regularized using 0.2 dropout and 0.01 weight decay coefficient. We run the experiments 5 times with different initialization and report the mean and standard deviations.

\section{Results}
We use Area Under Precision-Recall (AUCPR) metric for in-hospital mortality and decompensation tasks as they suffer from class imbalance with only 10\% patients suffering mortality, following the benchmark.~\newcite{davis2006relationship} suggest  AUCPR for imbalanced class problems. We use Cohen's linear weighted kappa, which measures the correlation between predicted and actual multi-class buckets to evaluate LOS in accordance with with~\newcite{harutyunyan2017multitask}. 

We compared the multimodal network with the baseline time series LSTM models for all three tasks. Results from our experiments are documented in Table \ref{tab:results}. Our proposed multimodal network outperforms the time series models for all three tasks. For in-hospital mortality prediction, we see an improvement of around 7.8\% over the baseline time series LSTM model. The other two problems were more challenging itself than the first task, and modeling the notes for the sequential task was difficult. With our multimodal network, we saw an improvement of around 6\% and 3.5\% for decompensation and LOS, respectively.  

We did not observe a change in performance with respect to results reported in benchmark~\cite{harutyunyan2017multitask} study despite dropping patients with no notes or chart time. In order to understand the predictive power of clinical notes, we also train text only models using CNN part from our proposed model. Additionally, we try average word embedding without CNN as another method to extract feature from the text as a baseline. Text-only-models perform poorly compared to time-series baseline. Hence, text can only provide additional predictive power on top of time-series data.

\begin{table}[t]
	\setlength{\textfloatsep}{0.1cm}
	\begin{tabular}{|c c c c c c|} 
		\multicolumn{6}{c}{\textbf{In-Hospital Mortality}}\\ [0.5ex] 
		\hline
		\multicolumn{2}{|c|}{} & \multicolumn{2}{c|}{AUCROC} & \multicolumn{2}{c|}{AUCPR} \\
		\hline
		\multicolumn{2}{|c|}{Baseline (No Text)} & \multicolumn{2}{c|}{0.844} & \multicolumn{2}{c|}{0.487}  \\
		\hline
		\multicolumn{2}{|c|}{Text-Only} &
		\multicolumn{2}{c|}{0.793} & \multicolumn{2}{c|}{0.303}  \\
		\hline
		\multicolumn{2}{|c|}{MultiModal - Avg WE} &
		\multicolumn{2}{c|}{0.851} & \multicolumn{2}{c|}{0.492}  \\
		\hline
		\multicolumn{2}{|c|}{MultiModal - 1DCNN} &
		\multicolumn{2}{c|}{\textbf{0.865}} & \multicolumn{2}{c|}{\textbf{0.525}}  \\
		\hline
		\multicolumn{6}{c}{}\\
		\multicolumn{6}{c}{\textbf{Decompensation}}\\ [0.5ex] 
		\hline
		\multicolumn{2}{|c|}{} & \multicolumn{2}{c|}{AUCROC} & \multicolumn{2}{c|}{AUCPR} \\
		\hline
		\multicolumn{2}{|c|}{Baseline (No Text)} & \multicolumn{2}{c|}{0.892} & \multicolumn{2}{c|}{0.325}  \\
		\hline
		\multicolumn{2}{|c|}{Text-Only} & \multicolumn{2}{c|}{0.789} & \multicolumn{2}{c|}{0.081}  \\
		\hline
		\multicolumn{2}{|c|}{MultiModal - Avg WE} &
		\multicolumn{2}{c|}{0.902} & \multicolumn{2}{c|}{0.311}  \\
		\hline
		\multicolumn{2}{|c|}{MultiModal - 1DCNN} &
		\multicolumn{2}{c|}{\textbf{0.907}} & \multicolumn{2}{c|}{\textbf{0.345}}  \\
		\hline
		\multicolumn{6}{c}{}\\
		\multicolumn{6}{c}{\textbf{Length of Stay}}\\ [0.5ex] 
		\hline
		\multicolumn{2}{|c|}{} & \multicolumn{4}{c|}{Kappa}\\
		\hline
		\multicolumn{2}{|c|}{Baseline (No Text)} & \multicolumn{4}{c|}{0.438} \\
		\hline
		\multicolumn{2}{|c|}{Text Only} & \multicolumn{4}{c|}{0.341} \\
		\hline
		\multicolumn{2}{|c|}{MultiModal - Avg WE} &
		\multicolumn{4}{c|}{0.449}  \\
		\hline
		\multicolumn{2}{|c|}{MultiModal - 1DCNN} & \multicolumn{4}{c|}{\textbf{0.453}}   \\
		\hline
	\end{tabular}
	\caption{Mean metrics for all three tasks. Standard deviations: IHM (AUCROC $<$ 0.004, AUCPR $<$ 0.015), Decompensation (AUCROC $<$ 0.008, AUCPR $<$ 0.008), and LOS (Kappa $<$ 0.003).}
	\label{tab:results}
	\vspace{-1.5em}
\end{table}
\vspace{-0.3em}

\section{Conclusion}
Identifying the patient's condition in advance is of critical importance for acute care and ICU management. Literature has exclusively focused on using time-series measurements from ICU instruments to this end. In this work, we demonstrate that utilizing clinical notes along with time-series data can improve the prediction performance significantly. In the future, we expect to improve more using advanced models for the clinical notes since text summarizes expert knowledge about a patient's condition.

\section{Acknowledgments}
We would like to thank Google Cloud for a generous funding for the computing resources used for this work.

\bibliography{emnlp-ijcnlp-2019}
\bibliographystyle{acl_natbib}
\end{document}